\documentclass{article}

\usepackage{microtype}
\usepackage{graphicx}
\usepackage{subfigure}
\usepackage{booktabs} 

\usepackage{hyperref}

\usepackage[accepted]{icml2023}

\usepackage{amsmath}
\usepackage{amssymb}
\usepackage{mathtools}
\usepackage{amsthm}
\usepackage{multirow}
\usepackage[capitalize,noabbrev]{cleveref}

\theoremstyle{plain}

\theoremstyle{definition}

\theoremstyle{remark}

\usepackage[textsize=tiny]{todonotes}

\icmltitlerunning{Personalized Prediction of Recurrent Stress Events Using Self-Supervised Learning on Multimodal Time-Series Data}

\begin{document}

\twocolumn[
\icmltitle{Personalized Prediction of Recurrent Stress Events Using Self-Supervised Learning on Multimodal Time-Series Data}



\icmlsetsymbol{equal}{*}

\begin{icmlauthorlist}
\icmlauthor{Tanvir Islam}{yyy}
\icmlauthor{Peter Washington}{yyy}


\end{icmlauthorlist}

\icmlaffiliation{yyy}{Information and Computer Sciences Department, University of Hawaii at Manoa, Honolulu, Hawaii, USA}

\icmlcorrespondingauthor{Tanvir Islam}{tislam@hawaii.edu}

\icmlkeywords{Machine Learning, ICML}

\vskip 0.3in
]



\printAffiliationsAndNotice{} 

\begin{abstract}

Chronic stress can significantly affect physical and mental health. The advent of wearable technology allows for the tracking of physiological signals, potentially leading to innovative stress prediction and intervention methods. However, challenges such as label scarcity and data heterogeneity render stress prediction difficult in practice. To counter these issues, we have developed a multimodal personalized stress prediction system using wearable biosignal data. We employ self-supervised learning (SSL) to pre-train the models on each subject's data, allowing the models to learn the baseline dynamics of the participant's biosignals prior to fine-tuning the stress prediction task. We test our model on the Wearable Stress and Affect Detection (WESAD) dataset, demonstrating that our SSL models outperform non-SSL models while utilizing less than 5\% of the annotations. These results suggest that our approach can personalize stress prediction to each user with minimal annotations. This paradigm has the potential to enable personalized prediction of a variety of recurring health events using complex multimodal data streams.

\end{abstract}

\section{Introduction}

Chronic stress can have profound detrimental effects on health and well-being. Research has shown that prolonged exposure to stress hormones can lead to increased risk of mental health disorders such as anxiety and depression \cite{mcewen2008central} and can contribute to the development of cardiovascular diseases, including hypertension and heart disease \cite{chrousos1992concepts}. Furthermore, chronic stress has been associated with dysregulation of the immune system, leading to impaired immune function and increased susceptibility to infections and autoimmune disorders \cite{dhabhar2014effects}. 

Biosignal-based stress detection methods are of increasing interest to the digital health community due to their potential to recognize stress levels in real-time. Commonly used biosignals for stress prediction include electrocardiograms (ECG) \cite{rabbani2022contrastive}, galvanic skin response (GSR) \cite{Lundberg1994}, electromyograms (EMG) \cite{inproceedingsstresssss, CRESCENTINI2016304, inbookmentalstress}, skin temperature (ST) \cite{lorig_2016}, skin conductivity \cite{7323251}, and respiratory rate  \cite{7838780}.

A fundamental challenge for the prediction of stress from multimodal biosignals is the engineering of appropriate features. Traditional machine learning approaches face a significant challenge in need for manual feature generation, and the feature representation approaches vary depending on the physiological signal \cite{articlestresslimitations, washington2022challenges}. These manually created features, although widely used \cite{7323251, prevfeat}, often do not lead to optimal performance outcomes for subjective prediction targets such as stress due to the inherent subjectivity, complexity, and heterogeneity of the data. Deep neural networks (DNNs) have revolutionized machine learning \cite{lecun2015deep} by learning to extract complex data patterns across various fields \cite{pmlr,74}, including in the medical sciences \cite{10.1001/jama.2016.17216}, without the need for manual feature extraction \cite{articlestresslimitations, Hannun2019}.

Another major challenge in machine learning with multimodal biosignal data is obtaining datasets with high-quality labels \cite{washington2023review, washington2023crowdsourcing}, a process which is notably costly and where the clinical ground truth is often difficult to define \cite{kalantarian2019labeling, washington2020data}. Annotation of stress events is laborious for the end user, leading to relatively sparse labels in most, if not all, datasets containing biosignals. The sparse labels, coupled with the user-dependent nature of the physiological stress response, lead to immense difficulty in training a generalizable stress prediction model.

To address these two challenges, which have traditionally led to a subpar performance in the use of machine learning prediction of subjective psychiatric outcomes such as stress, we suggest a personalized self-supervised learning (SSL) approach to stress prediction. By training a separate model per user (personalization) and using SSL to learn the baseline dynamics of each user's biosignals, we are able to learn using relatively few annotations from each user, even for annotations which are inherently subjective such as stress. We capitalize on multiple sensor modalities, performing stress recognition through the integration of diverse information \cite{Walambe2021}. We use electrodermal activity (EDA), electrocardiogram (ECG), electromyography (EMG), respiration (RESP), core body temperature (TEMP), and three-axis acceleration (ACC) to train our stress prediction models. 

Our contributions to the machine learning for the healthcare field are as follows:

\begin{itemize}
    \item We evaluate stress prediction models trained using multiple biosignal modalities.
    \item We explore the personalization of these multimodal stress prediction models, enabled through SSL procedures which learn the baseline temporal dynamics of each user's biosignals.
    \item We compare multimodal personalized models with and without SSL pre-training to quantify the impact of personalized multimodal SSL.
\end{itemize}

\section{Related Works}
\label{related work}

Stress prediction from biosignals is a rich and expanding field. Established machine learning approaches have yielded preliminary successes while leaving room for improvement. Fern{\'a}ndez et al. introduced a non-invasive, radar-based stress detection method that primarily uses respiratory patterns and applied Recurrence Quantification Analysis (RQA) to achieve 94.4\% accuracy, offering broader applicability than heartbeat-based methods, particularly for individuals with a higher body mass index \cite{fernandez2018mental}. Ghaderi et al. employed machine learning algorithms to process various biological signals for accurate stress level classification, emphasizing the role of respiration as a crucial sensor and suggesting its potential applications for tailoring stress management treatments \cite{ghaderi2015machine}. Karthikeyan et al. studied the correlation between stress levels and EMG measurements of muscle tension \cite{10.1007/978-3-642-35197-6_26}. Other lines of research emphasize the utility of ECG data in stress measurement, including a novel ECG-based mono-fuzzy measure to assess stress levels \cite{articlejonwun} as well as the application of person-specific and person-independent stress prediction methods \cite{9734747}.

SSL has only recently been applied to biosignals-based stress prediction. Robert et al. proposed a stress prediction system based on contrastive learning, utilizing modified EDA signals to classify stress and non-stress scenarios \cite{matton2022contrastive}. Contrastive learning techniques such as SimCLR \cite{cheng2020subject} and BOYL \cite{grill2020bootstrap} heavily rely on data augmentation. However, this approach has limitations, as determining the most beneficial augmentations is challenging \cite{zhang2022rethinking}. By contrast, our study explores a different approach by excluding data augmentation and instead focusing on self-supervised pre-training using the raw data. 

Multimodal machine learning techniques are required to handle the multiple concurrent biosignals which are recorded by consumer and research wearables. Bobade et al. detected stress levels in individuals by analyzing various bio-signals such as acceleration, ECG, blood volume pulse, body temperature, respiration, EMG, and EDA. Their experimental results included accuracy levels of up to 84.32\% for three-way classification (amusement vs. baseline vs. stress) and up to 95.21\% accuracy for binary classification (stress vs. non-stress), outperforming previous work in the field \cite{bobade2020stress}. 
Aigrain et al. developed a methodology for analyzing multimodal stress detection results by considering multiple assessments of stress \cite{aigrain2016multimodal}. Data from 25 subjects in a stressful situation were collected, along with three different assessments: self-assessment, assessments from external observers, and assessments from a physiology expert. The study found that a combination of behavioral and physiological features, such as body movement, blood volume pulse, and heart rate, provided valuable information for classifying stress across the three assessments \cite{aigrain2016multimodal}. Bara et al. introduced a deep learning-based approach for multimodal stress detection, utilizing the MuSE dataset \cite{inproceedingsmuse} to evaluate various configurations and modular architectures. The results demonstrated the potential of deep learning methods in capturing affective representations related to stress, paving the way for further investigations and applications in affective computing \cite{bara2020deep}. 

In contrast to these prior works, we combine the fields of multimodal machine learning and SSL to achieve optimal performance \textit{with a minimal number of training labels}. Multimodal models traditionally require more labels to accommodate the expanded feature space, but we hypothesize that SSL can drastically reduce this need. Furthermore, we combine these ideas with the idea of \textit{model personalization}, where a separate model is trained per individual. 

\section{Methodology}

\label{Methodology}

\begin{figure*}
    
    \includegraphics[width=\linewidth]{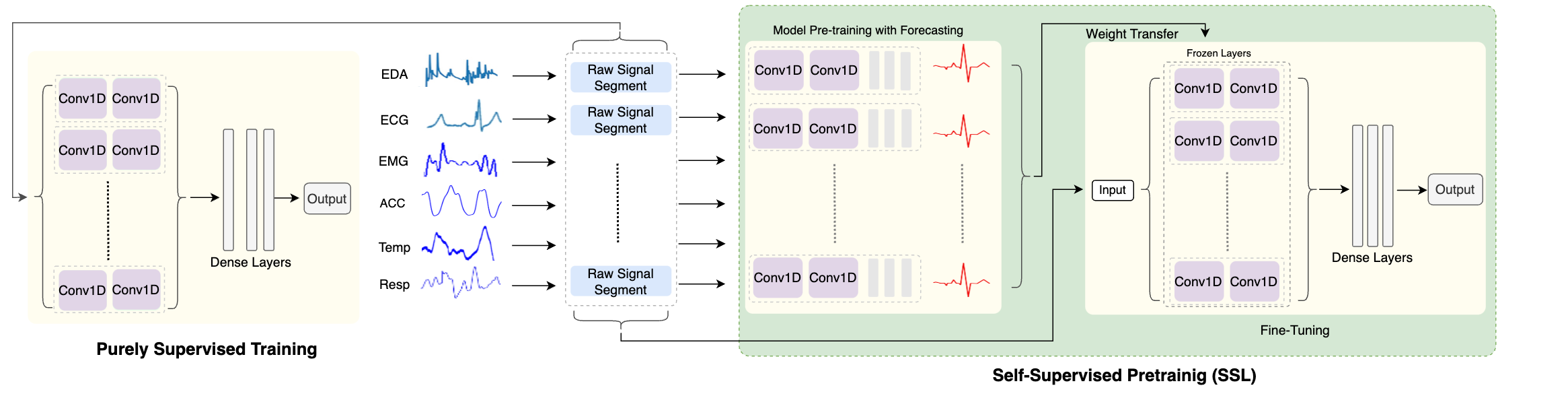}
    \caption{Personalized stress prediction framework involves segmenting the raw signal into overlapping data points and utilizing a self-supervised pre-training forecasting model (right model), which is then fine-tuned for stress prediction. We compare this model to a supervised learning model (left model) without self-supervised pre-training to assess its effectiveness.}
    \label{overall architecture}
\end{figure*}

We aim to develop a model that can accurately predict stress by learning representations of six concurrent biosignals: EDA, ECG, EMG, Temp, Resp, and ACC. Our model is developed in two stages: (1) pre-training to predict a pretext task $T_p$ and (2) fine-tuning to the downstream task $T_d$, which in this case is stress prediction. In the pretext task ($T_p$), the goal is to learn robust and generalized features from unlabeled biosignals through a self-supervised process which learns meaningful representations that can capture relevant patterns and information from each user's individual biosignal dynamics. Once the representations are learned from the pretext task, the second stage ($T_d$) focuses on predicting stress using the original biosignal data along with human-annotated stress labels ($y_i$). We compare the fine-tuned SSL-based model against a purely supervised model. For fully supervised training, we use the same architecture as $T_d$. 
Figure~\ref{overall architecture} illustrates the proposed self-supervised solution, depicting the two-stage process of learning representations and predicting stress.

\subsection{Dataset}

We use a publicly available dataset called WESAD \cite{schmidt2018introducing}, consisting of data collected using the Empatica E4 wrist-worn device and RespiBAN device worn on the chest. WESAD contains six biosignals per participant: ECG, EDA, EMG, RESP, TEMP, and ACC. In addition to the physiological data collected, participants periodically completed questionnaires assessing their emotional state during the data collection session. Our models are trained exclusively on the RespiBAN device data.

During the data collection process, information related to three distinct affective states, namely stress, amusement, and relaxation, were gathered from a group of 15 participants. In order to assess each participant's anxiety levels, they were presented with six questions derived from the State-Trait Anxiety Inventory (STAI). These questions required the participants to indicate their responses on a four-point Likert scale. The six STAI questions covered six distinct affective states: \textit{feeling at ease, nervous, jittery, relaxed, worried, and pleasant}. By asking these STAI questions, the developers of WESAD aimed to gauge each participant's anxiety levels and to capture their subjective emotional states.

Our models were trained using these six biosignals from the RespiBAN device for baseline condition, thus facilitating a multimodal learning training process.

\subsection{Label Representation}
The WESAD dataset includes participants' responses to six questions from the STAI, which were rated on a four-point Likert scale (ranging from 1 to 4). In this paper, we address the task of stress detection by proposing a label quantification approach. Rather than treating stress detection as a discrete classification problem with predefined stress levels, we transform it into a regression problem by assigning continuous values to the stress labels. Specifically, we convert the stress labels of 1, 2, 3, and 4 into quantified values of 0.25, 0.5, 0.75, and 1, respectively. To ensure a quantified and uniformly distributed representation of labels, we perform a conversion based on the evenly spaced four-point Likert scale. This label quantification technique allows us to capture the underlying intensity or severity of stress, enabling more fine-grained analysis and prediction. By leveraging DNN over raw biosignals, we can now model and predict stress levels with greater precision and sensitivity, contributing to more accurate and nuanced stress prediction systems. Our experimental results demonstrate the effectiveness of label quantification in improving the performance and interpretability of stress prediction models, highlighting its potential for advancing the field of stress analysis and management.

\subsection{Self-supervised Learning of Biosignal Representations}

Self-supervised pre-training involves gaining a comprehensive understanding of data without relying on labeled information, known as the ``pretext task''. By training a model with a robust representation, it becomes more feasible to transfer to a specific task. We trained separate models for each subject using only their respective data, allowing the models to learn the baseline dynamics of each individual.

\begin{table}
\label{pretrainNetwork}
\caption{Architectural details of the pre-trained network (1D CNN).}
\begin{tabular}{|p{1cm}|p{2.5cm}|p{1.5cm}|p{1.5cm}|}

\hline Module & \multicolumn{1}{c|}{Layer Details} & Activation & Shape \\
\hline \hline \small Input & \multicolumn{1}{c|}{-} & \multicolumn{1}{c|}{-} & \small $7000 \times 1$ \\
\hline \multirow{10}{*}{ \centering \small Layers } & \small { Conv1D, four $1\times40$ kernels } &\small Leaky-ReLU& \small $7000 \times 40$ \\
\cline{2-4}
& \small {Conv1D, two $1\times30$ kernels } & \small Leaky ReLU& \small $7000 \times 30$ \\
\cline{2-4}
& \small {Conv1D, four $1\times18$ kernels}& \small Leaky-ReLU& \small $7000 \times 18$ \\
\cline{2-4}
& \small {Conv1D, two $1\times30$ kernels } &  \small Leaky-ReLU& \small $7000 \times 30$ \\
\cline{2-4}
& \multicolumn{1}{c|}{Dense} &\small Leaky-ReLU& \small $1 \times 70$ \\
\cline{2-4}
& \multicolumn{1}{c|}{Dense} &\small Leaky-ReLU& \small $1 \times 30$ \\
\hline \small Output & \multicolumn{1}{c|}{-}  &\small Linear&\small  40 \\
\hline
\end{tabular}
\label{pretrainNetwork}
\end{table}

The model is pre-trained by dividing each of the six signals into fixed windows of $7000$ ($10$ seconds) dimensions, resulting in $7910$ data points with a 100-point overlap using the forecasting method. Each training data point consists of a $7000$-dimensional vector, with the target output being the subsequent $40$ data points. We set the window size to $10$ seconds because we aim to predict stress for $10$ seconds. A 1D CNN is utilized for the pre-training process. The 1D CNN architecture of this model pre-training is presented in Table~\ref{pretrainNetwork}.

Let the original signal sequence is $S = {S_1, S_2, ..., S_n}$. We then create a series of training data points $X$ and corresponding targets $Y$ as follows: For each $i$ from $1$ to $D$ (where $D$ is the total number of data points), we have $ X_{i} = {S_{(i-1)*O + 1}, ..., S_{(i-1)*O + W}}$ and the corresponding target $Y_{i} = {S_{(i-1)*O + W + 1}, ..., S_{(i-1)*O + W + P}}$, where $S$ is the original signal, $D$ is the total number of data points, $W$ is the window size (7000 in our case),  $P$ is the number of points to predict (40 in our case), $O$ is the overlap (100 in our case), $X_i$ is the $i$-th data point (an array of $W$ elements from the original signal), $Y_i$ is the corresponding target (an array of $P$ elements from the signal, immediately following the elements in $X_{i}$). 

Each $X_{i}$ is a window of the original signal, and each $Y_{i}$ is the sequence of points immediately following the corresponding $X_{i}$ in the signal.

With this method, we have learned baseline representations of all biosignal modalities \textit{for each user}: $R_{EDA}$, $R_{ECG}$, $R_{EMG}$, $R_{Temp}$, $R_{ACC}$, $R_{Resp}$ 

\subsection{Stress Prediction Task}
\begin{figure*}
    \includegraphics[width=\linewidth]{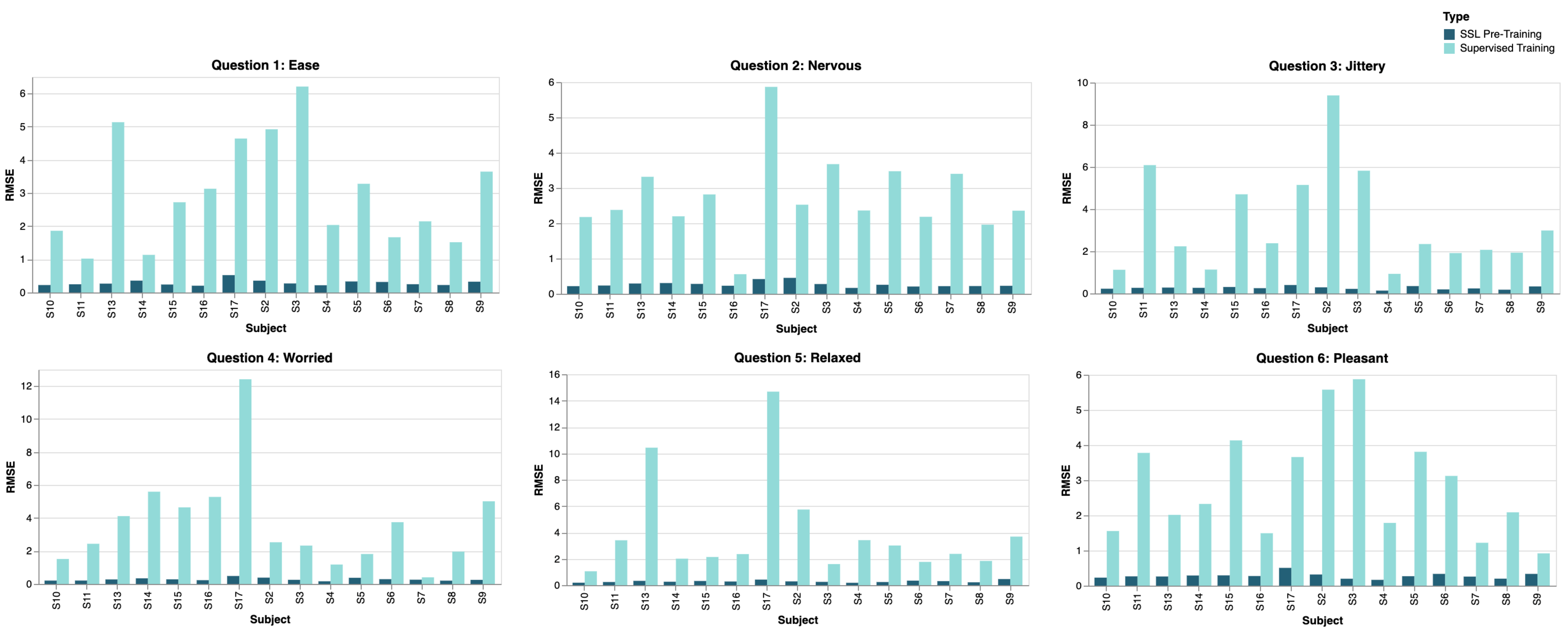}
    \caption{RMSE between a model fine-tuned after SSL and a model trained purely with supervised training for all subjects and all annotation types.}
    \label{evaluationallsubs}
\end{figure*}

In the second phase of learning ($T_d$), the model uses the initial biosignals and responses provided by participants ($y_i$). In this phase, the framework utilizes the fixed convolution layers from the first network to extract features that will help to predict stress. The fully-connected layers at the end of the network are subsequently trained in a supervised manner to predict stress using the extracted features. 
\begin{table}
\caption{Architectural details of the fine-tuned network (1D CNN).}
\begin{tabular}{|p{1cm}|p{2.5cm}|p{1.5cm}|p{1.5cm}|}

\hline Module & Layer Details & Activation & Shape \\
\hline \hline Input & \multicolumn{1}{c|}{-} & \multicolumn{1}{c|}{-} & $7000 \times 1$ \\
\hline \small Frozen Layers  & \small {Conv1D, four $1\times40$ kernels} &\small Leaky ReLU& $\small 7000 \times 40$ \\
\cline{2-4}

& \small {Conv1D, two $1\times30$ kernels} & \small Leaky ReLU& \small $7000 \times 30$ \\
\cline{2-4}

& \small {Conv1D, four $1\times18$ kernels} &\small Leaky ReLU& \small $7000 \times 18$ \\
\cline{2-4}

& \small {Conv1D, two $1\times30$ kernels } & \small Leaky ReLU & \small $7000 \times 30$ \\
\cline{2-4}

\hline \small Task-Specific Layers & \multicolumn{1}{c|} { Dense } &\small Leaky ReLU& \small $1 \times 50$ \\
\cline{2-4}

& \multicolumn{1}{c|} { Dense } &\small Leaky ReLU&\small  $1 \times 30$ \\
\cline{2-4}

& \multicolumn{1}{c|} { Dense } &\small Leaky ReLU&\small  $1 \times 30$ \\
\hline Output & \multicolumn{1}{c|}{-} &\small Linear& \small 1 \\
\hline
\end{tabular}
\label{FinetuneNetwork}
\end{table}
In order to perform the stress prediction task ($T_d$) using multiple modalities, the representations of the biosignals, namely $R_{EDA}$, $R_{ECG}$, $R_{EMG}$, $R_{Temp}$, $R_{ACC}$, and $R_{Resp}$, are fused together into a single multimodal representation $R$. This late-stage fusion process combines the information from each modality to create a comprehensive representation that captures the relevant features for predicting stress. To accomplish $T_d$ with representation $R$, a network: $\rho = w(R, \theta)$ is trained where $\rho$ is the output prediction vector and $\theta$ is
the set of trainable parameters. Finally, we calculate the optimal value of $\theta$ by minimizing mean squared error loss:
$$\text{L} = \frac{1}{n} \sum_{i=1}^{n} (y_i - \hat{y}_i)^2
$$
where $n$ represents the number of data points, $y_i$ represents the observed values, and $\hat{y}_i$ represents the predicted values.

We describe the details of the network that is used for the downstream stress prediction task ($T_d$) in Table~\ref{FinetuneNetwork}.

\subsection{Experimental Procedures}

We examine the performance of the models pre-trained with SSL in comparison to models trained solely through supervised training without pre-training. To conduct this comparison, we utilize a dataset consisting of $7,000$ data points from six biosignals, with each data point representing a $10$-second window and all six signal modalities. These data points are used to train both types of models (SSL pre-training followed by supervised fine-tuning and purely supervised training). To test the models, we hold out the last set of $910$ data points with $10$-second windows. Both models are evaluated using the same test set for each subject. 

\begin{figure*}
    \includegraphics[width=\linewidth]{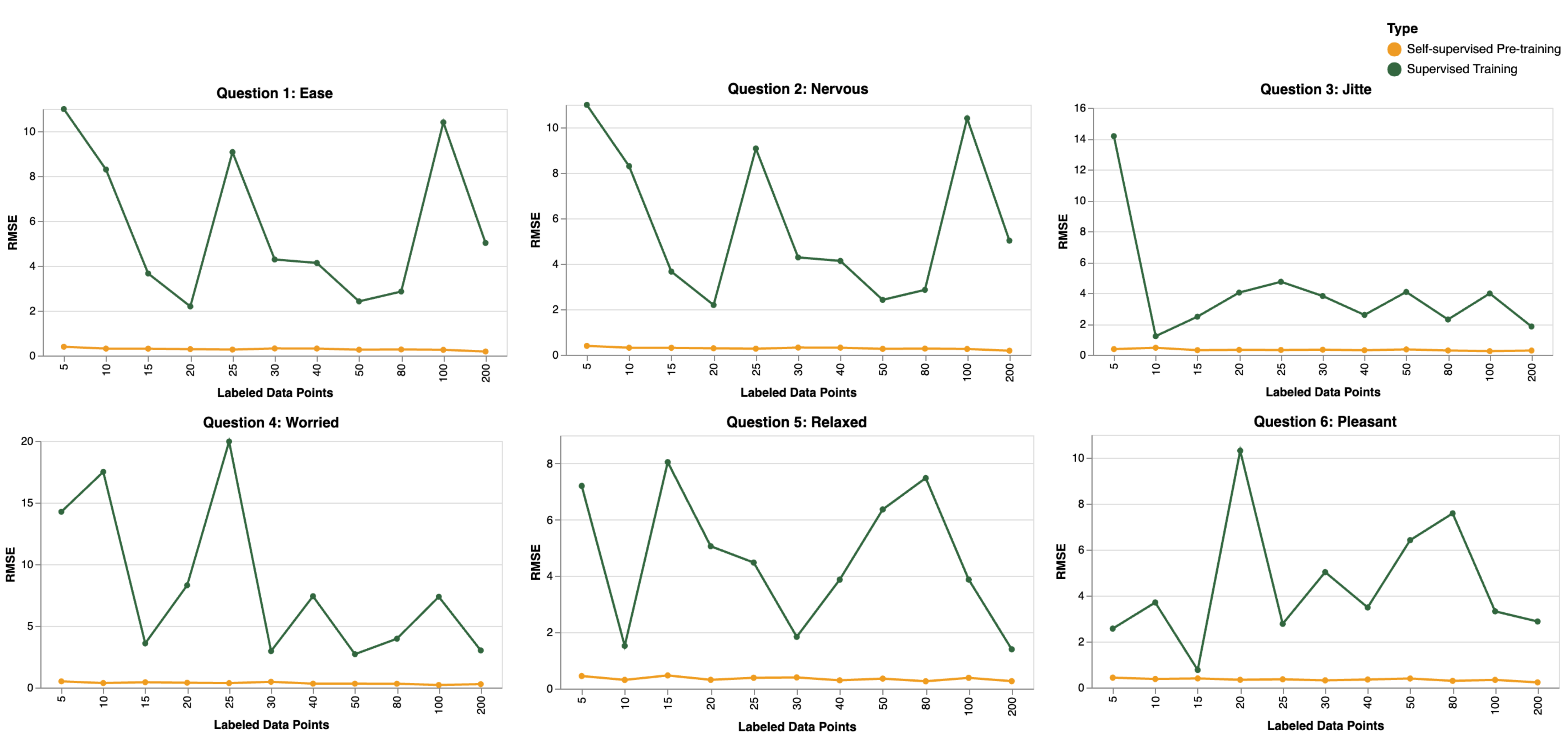}
    \caption{Comparison of the performance between SSL pre-training and solely supervised training methods for a demonstrative user (Subject 2). Similar trends were observed for all subjects. }
    \label{diffdatapoints}
\end{figure*}
\section{Results and Evaluation}

We evaluate each model using RMSE (Root Mean Squared Error). We show the RMSE scores for all subjects across various questions in Figure~\ref{evaluationallsubs}. The scores were obtained by training 10 sets of random data points three times, each time using different random data samples, and calculating the average RMSE score. 
We observe that self-supervised pre-training achieves significantly lower RMSE compared to models trained solely through supervised learning. This indicates that SSL, which involves learning representations from unlabeled data, results in improved performance and reduced prediction errors compared to models trained only with labeled data. The figure demonstrates the superiority of SSL pre-training in terms of RMSE scores.

We also compare the models pre-trained with SSL and the purely supervised model using a set of different labeled data points. Figure~\ref{diffdatapoints} shows the comparison between these two models for Subject 2 for all questions. Figure~\ref{diffdatapoints} demonstrates that even when sampling only 5 random data points, the SSL pre-trained model repeatedly outperforms purely supervised training for all users and all outcome measures.

\section{Discussion}

We present an analysis of a personalized multimodal learning framework, focusing on the advantages of self-supervised pre-training in comparison to a fully-supervised training paradigm. The results of this study emphasize the superiority of SSL over purely supervised training methods within the context of personalized learning. We observe this benefit with only a few labeled examples, demonstrating the personalized SSL can enable learning of complex and subjective outcomes such as stress using relatively small datasets.

With the aid of SSL, the personalized models can leverage both labeled and unlabeled data, making more efficient use of available information. This is especially valuable in real-world situations where limited labeled data or data of varying quality is present such as in the personalization of machine learning-powered healthcare applications \cite{daniels2018feasibility, daniels2018exploratory, kline2019superpower, voss2019effect, washington2017superpowerglass}, where ample data are recorded per individual but each label is burdensome for the end-user to provide.  By contrast, purely supervised models display high variability in their performance as indicated by a significant fluctuation in RMSE values across different training runs or data samples. This variability can be problematic for personalized applications, where consistent and dependable outcomes are crucial.

There are some limitations of this study which should be addressed in follow-up work. The WESAD dataset was created in a highly controlled environment. Future research endeavors should involve comprehensive and unconstrained data collection setups to test the system using real-time, ``in-the-wild'' data streams. Such experimental procedures will provide valuable insights into multimodal personalization in real-world scenarios. 

\section{Conclusion}

We introduce a novel approach for enhancing deep learning models to predict stress using multimodal biosignals. Our technique empowers deep learning models to adapt to an individual's unique baseline temporal dynamics, enabling more precise and personalized predictions of stress with significantly fewer annotations required. By leveraging multi-modal biosignals, our method opens up new possibilities for understanding and addressing stress-related challenges in various contexts, including healthcare, workplace environments, and personalized mental health interventions. The potential impact of our findings is substantial, as the personalized learning technique presented here significantly reduces the need for extensive human annotation typically associated with deep learning models for healthcare.

\section*{Acknowledgements}

The technical support and advanced computing resources from University of Hawaii Information Technology Services – Cyberinfrastructure, funded in part by the National Science Foundation CC* awards \# 2201428 and \# 2232862 are gratefully acknowledged.

\nocite{langley00}

\bibliography{paper}
\bibliographystyle{icml2023}

\newpage

\end{document}